\documentclass[11pt,a4paper]{article}
\usepackage[hyperref]{emnlp2020}
\usepackage{times}
\usepackage{latexsym}
\usepackage[utf8]{inputenc}
\usepackage{csquotes}

\usepackage{microtype}

\aclfinalcopy % Uncomment this line for the final submission
\def\aclpaperid{3068} %  Enter the acl Paper ID here

\usepackage[utf8]{inputenc}
\usepackage{enumitem}
\setlist{nosep}
\usepackage{linguex}
\usepackage{url}
\usepackage{latexsym}
\usepackage{booktabs}
\usepackage{multirow}
\usepackage{graphicx}
\usepackage{float}
\usepackage{csquotes}
\usepackage{colortbl}
\usepackage{amssymb}
\usepackage{amsmath}
\usepackage{bm}
\usepackage{rotating}
\usepackage{diagbox}
\usepackage{stfloats}
\usepackage{scalerel}
\usepackage{soul, color}
\usepackage{comment}
\usepackage{blindtext}

\newcommand\blfootnote[1]{%
  \begingroup
  \renewcommand\thefootnote{}\footnote{#1}%
  \addtocounter{footnote}{-1}%
  \endgroup
}

\title{Help! Need Advice on Identifying Advice}

\author{Venkata S Govindarajan$^1$\ \ \ \ 
Benjamin T Chen$^{2*}$\ \ \ \ 
Rebecca Warholic$^{3\dagger}$\ \ \ \ \\
\textbf{Katrin Erk}$^1$\ \ \ \ 
\textbf{Junyi Jessy Li}$^1$\\
$^1$ Department of Linguistics, The University of Texas at Austin\\
$^2$ Amazon Inc.\\
$^3$ McGill University\\

{\tt \{venkatasg,btchen\}@utexas.edu,
rebecca.warholic@mail.mcgill.ca,}\\
{\tt katrin.erk@utexas.edu, jessy@austin.utexas.edu}
}

\date{}

\begin{document}
\maketitle\blfootnote{* Work done as an undergraduate student at UT Austin. \\$\dagger$ Work done at UT Austin while on the DREU undergraduate research program.}

\begin{abstract}
Humans use language to accomplish a wide variety of tasks --  asking for and giving advice being one of them. In online  advice  forums, advice is mixed in with non-advice, like emotional support, and is sometimes stated explicitly, sometimes implicitly. Understanding the language of advice would equip systems with a better grasp of language pragmatics; practically, the ability to identify advice would drastically increase the efficiency of advice-seeking online, as well as advice-giving in natural language generation systems.

We present a dataset in English from two Reddit advice forums -- r/AskParents and r/needadvice -- annotated for whether sentences in posts contain advice or not. Our analysis reveals rich linguistic phenomena in advice discourse. We present  preliminary models showing that while pre-trained language models are able to capture advice better than rule-based systems, advice identification is challenging, and we identify directions for future research.

\end{abstract}

\setlength{\Exlabelsep}{0em}
\setlength{\Extopsep}{.2\baselineskip}
\setlength{\SubExleftmargin}{1.3em}

\section{Introduction}
\label{sec:introduction}
Humans use language in the real world to achieve many goals -- communicate intents and desires, to argue and convince, and to ask for and give advice. In recent years, people have increasingly looked to the internet to find advice; advice forums like \href{https://www.babycenter.com}{BabyCenter} and \href{https://reddit.com/r/needadvice/}{r/needadvice} have hundreds of thousands of members; studies also showed that people increasingly seek health advice online~\cite{pewstudy,chen2018health}. However, finding the right  solution to a problem is difficult, since advice may be spread over multiple posts and pages online. Even within the same post, not all sentences contain relevant advice, like in the following (truncated) reply to a question titled \textit{Is it too late to start a hobby/activity at 12?}:

\ex. \textit{..you can always pick anything up you think is interesting and giving it a shot}. You never know what you are good at until you try new things! \textit{Idk if you have a budget or maybe borrow tools but you can try woodworking?} It's fun and frustrating (in a good way) at the same time \label{ex:advice-1}

Only the italicised sentences are advice to the question asked. Both sentences that follow the advice sentences lend support to the advice, rather than containing advice towards a course of action themselves. People also give advice in different ways~\cite{abolfathiasl2013pragmatic}, often implicitly like in the following reply to a question titled \textit{Parenting with a history of depression?}, where advice is implicitly conveyed via personal experience:

\ex. I took my meds the whole time. I used the tools I learned in therapy. I talked on Reddit with others to get support and ideas. \label{ex:advice-2}

Automatic identification of advice in text would thus be extremely useful. Yet, as we see above, it would also require a deep understanding of semantics and discourse pragmatics. In recent years, NLP systems based on large-scale pre-trained language models have shown impressive gains on several linguistic benchmarks \citep{devlin_bert:_2019, Clark2020ELECTRA, yang_xlnet_2019}. However, these same models have been found to struggle at tasks that require higher-level processing~\cite{ettinger2020bert}, including giving advice \citep{zellers2020evaluating}.

This work aims to advance both our understanding of how people give advice, as well as to provide resources for learning to identify advice. First, we construct a dataset of annotations of advice in English from two advice-focused Reddit communities -- \href{https://old.reddit.com/r/AskParents/}{r/AskParents} and \href{https://old.reddit.com/r/needadvice/}{r/needadvice}, totalling 18456 sentences across 684 posts (\S\ref{sec:datacollection}). These two subreddits are different in a number of respects. r/needadvice is a general advice forum, while r/AskParents targets a specific audience---parents---who are often active seekers of advice. r/needadvice is more strongly moderated than r/AskParents. In addition, our analysis shows that r/AskParents contains more implicit, narrative advice than r/needadvice (\S\ref{sec:prelimanalysis}). 
Through this dataset we provide the first-of-its-kind resources to explore the breadth of advice-giving strategies, and testbeds for modeling advice.

We establish benchmarks for this task with BERT~\cite{devlin_bert:_2019}, a large pre-trained language model, to identify sentences that constitute advice.
We find that it is substantially better than a rule-based approach (\S\ref{sec:models}). In an in-depth analysis, we find that BERT re-discovers some linguistic rules that have been previously proposed for identifying advice, but struggles with advice that is more implicit, for example in the form of a narrative, like in \ref{ex:advice-2} (\S\ref{sec:analysis}). Our results also show that r/AskParents is more challenging for advice identification, despite the fact that r/needadvice has a wider range of topics. We make all of our data and code available online\footnote{\url{https://github.com/venkatasg/Advice-EMNLP2020}}.

\section{Related Work}
\label{sec:background}
\paragraph{Advice Strategies} There has been sociological and pragmatic work analysing how people navigate the task of engaging in advice discourse. People weigh interactional costs when giving and asking for advice \citep{shaw_managing_2013}, and they engage in various strategies to persuade their interlocutor and achieve their goals. Effective advice givers were found to engage in roles that extended beyond giving advice -- they help advice seekers clarify their problem, list possible solutions and sort through them, offer support and reassurance, and more \citep{decapua_strategies_1993}. While there has been work by \citet{fu-etal-2019-asking} looking at how people use personal narratives to \emph{ask for} advice online, no work thus far has looked at the discourse of advice \emph{giving} online.

\paragraph{SemEval} SemEval-2019 introduced a pilot task on suggestion mining \citep{negi-etal-2019-semeval},  recognizing the growing importance of identifying whether a text contains a suggestion towards a course of action or not. The dataset only considers sentences that \emph{explicitly } include suggestions -- that is, where one can infer without context that a sentence is a suggestion -- while we always give the annotators the wider context of the entire post and question, and ask them to evaluate which sentences are advice based on this wider context. For instance, \ref{ex:advice-2} is advice in the context of the question, but that same narrative could also be support for advice, given a different question. Additionally, suggestions are  not synonymous with advice, and can include tips and recommendations (although none of these terms are mutually  exclusive). For example, \emph{You should try the food at Italian restaurant} might be construed as a tip or a  recommendation, rather than advice.

SemEval-2019 Task 9 provides two datasets -- one from a  software suggestions forum and another from a hotel reviews website. While the dataset and the suggestion mining models are useful for understanding suggestions, we find that the definition of suggestion is too constrained -- explicit suggestions will not include many implicit instances of advice, which we are interested in studying. Secondly, we find the domain of their datasets to be somewhat restricted, and not representative of the wide range of online advice-seeking behavior. We chose to construct datasets based on subreddits devoted to asking for advice related to parenting and general issues, since we want to understand how to model general human advice-seeking interactions. We target parenting as parents frequently seek and give advice online, and express it in linguistically diverse forms. For general advice, r/needadvice has clear grouping mechanisms (``flairs'') that inform us with the topic of advice, which we use during analysis. 

\paragraph{TuringAdvice} Contemporaneous work from  \citet{zellers2020evaluating} introduces a new framework to evaluate the performance of language models. \texttt{TuringAdvice} challenges models to generate advice that is at least as helpful to the advice seeker as human generated advice. They introduce a new dataset called \textsc{redditadvice}, which scrapes posts from a wide  variety of advice subreddits. Annotators on Mechanical Turk were presented with a Reddit post seeking advice, along with two  replies to the post, and were asked to choose which reply constitutes the more helpful advice. 

However, as \ref{ex:advice-1} shows, the entirety of a response to a question rarely constitutes advice. In contrast, our work annotates and identifies explicit and implicit advice \emph{within} a reply to an advice-seeking posts and finds that less than 40\% of sentences in a reply are actually advice (Table~\ref{tab:dataset-sents}). Moreover, we focus on \emph{understanding} how people give advice linguistically, and to what extent pre-trained language models are able to identify advice. We believe our approach of  analyzing what constitutes advice at the semantic and  discourse level complements the motivation of~\citet{zellers2020evaluating}.

\section{Data Collection}
\label{sec:datacollection}
\subsection{Data sources}

In this section, we describe the data pipeline that we used to  collect annotations. We sourced our data from Reddit -- an online forum composed of many communities dedicated to specific topics (called subreddits). We gathered our data from two subreddits -- \href{https://old.reddit.com/r/AskParents/}{r/AskParents}, which is a forum for parents seeking advice on how to raise their children, and \href{https://old.reddit.com/r/needadvice/}{r/needadvice}, a general advice forum, where users (or moderators) also have the ability to tag their advice-seeking posts with a specific  flair (i.e. category). r/AskParents and r/needadvice were chosen for their respective narrow and wide domains (and audience), and also because we believed we might see differences in how advice is communicated based on our pilot studies. r/needadvice is also more highly moderated than r/AskParents, having more rules for users to follow for posting and replying to posts. We believe all of these factors contribute to two different ``styles'' of advice-giving.

For r/needadvice, we study posts which contain the following highly frequent flairs: ``Education", ``Career", ``Mental Health", ``Life Decisions", and ``Friendships". Some flairs were not considered due to the lack of variety in responses. For example, in the ``Medical" flair, replies often consisted of telling the original poster to see the doctor. 

\subsection{Annotation Task}

We crowdsource advice annotations from Amazon Mechanical Turk. Despite the inherent noise due to crowdsourcing~\cite{parde-nielsen-2017-finding}, recent work showed that when designed carefully, \emph{aggregated} crowdsourced annotations are trustworthy even for complex tasks~\cite{nye-etal-2018-corpus}.

As \ref{ex:advice-1} illustrates, not all sentences in a response to an advice-seeking question constitute advice. Thus, we want annotators to highlight which parts of the response to a question are advice, and which are not. We also want to find instances of implicit advice, i.e., advice that is given indirectly, like in \ref{ex:advice-2}. To ensure that annotators can also identify advice that might be marked using contextual cues, we provide annotators with sufficient context. 

In our task, we present annotators with an advice-seeking post and the post's corresponding replies. Given the hierarchical structure of forum replies, we show workers comment-trees, where a comment-tree is a comment and all of its replies\footnote{The order of comment-trees are determined by Reddit's ranking algorithm. We ordered by ``top'' comments}. Annotators are instructed (with examples) to highlight instances of both direct and indirect (implicit) advice within these comment trees. The highlighting interface, setup using the third-party tool \textsc{brat} \citep{stenetorp-etal-2012-brat}, asks annotators to highlight the longest contiguous span of 
text that they deem to be advice that addresses the 
question in the post.

\paragraph{Preprocessing} We recruited annotators on Amazon Mechanical Turk who were from the USA, had a minimum approval rating of 95\%, and had completed at least 500 HITS. To ensure that the posts on which annotators worked were substantive, we chose posts from both subreddits that were at least 3 days old and had at least 3 comments with 10 or more tokens. Comments made by the original poster or moderators usually did not contain any advice, so they were excluded\footnote{If the original poster makes a reply to an existing comment, we only annotate posts that appear \emph{before} that reply.}. To keep the task load reasonable for annotators, any posts with a submission title and body exceeding one standard deviation above the average length of posts (421 tokens) were filtered out; we restricted comment-trees to a depth of 2 and constructed HITS to contain at most 5 top-level comments to an advice-seeking post. Each HIT was annotated by 5 annotators for \$0.15 per HIT.  We perform a final round of preprocessing on our dataset to ensure quality~\cite{cachola2018expressively}, by removing annotations from workers whose Spearman correlation against the sum of labels within a HIT was below 0.2. 

\begin{table}[t]
    \centering
    \footnotesize
    \newcolumntype{L}{>{\centering\arraybackslash}m{1cm}}
\begin{tabular}{llll}
	\toprule
    \textbf{Dataset} & \textbf{Sentences} & $\bm{\kappa_{maj}}$ & $\bm{\kappa_{DS}}$  \\
    \midrule
    AskParents & 203 & 0.620 & 0.669 \\
    \midrule
    needadvice & 110 & 0.680 & 0.681 \\
    \bottomrule
\end{tabular}

    \caption{Gold annotator agreement on the internal task.}
    \label{tab:dataset-internal}
    \vspace{-4mm}
\end{table}

\subsection{Annotator agreement}

We use \emph{sentences} as our processing unit for advice identification. While \textsc{brat} does not restrict highlights to be along sentence boundaries, we observed that when a sentence contains highlights, 77.9\% of the tokens are highlighted, and that using sentences as units avoids fine-grained annotator variability resulting from the free-form highlighting interface.

\paragraph{Label aggregation} Following~\citet{nye-etal-2018-corpus}, we use the Dawid-Skene algorithm  \citep{dawid_maximum_1979}  to obtain aggregated labels, henceforth referred to as Dawid-Skene (DS) labels\footnote{We used \href{https://github.com/ipeirotis/Get-Another-Label}{Get-Another-Label} to generate DS labels}. This is an EM based algorithm that estimates the label with the maximum estimated posterior probability by iteratively computing annotator competencies and type probabilities. The algorithm ensures that competent annotators are given higher weight, and we show below that it is preferable to majority vote aggregation. 

\paragraph{Expert annotation} To evaluate the reliability of the DS labels, pilot annotations were done internally by three authors, two of whom are trained linguists. They also constructed an ``expert'' annotation of a randomly selected subset of posts, containing 203 sentences for r/AskParents and 110 sentences for r/needadvice. Cohen's Kappa~\citep{cohen_coefficient_1960}  was 0.529 for r/AskParents and 0.572 for r/needadvice, indicating moderate agreement. Disagreements in expert annotations were subsequently adjudicated to construct the gold annotations on the subset of posts.

\begin{table}[t]
    \centering
    \footnotesize
    \begin{tabular}{lllll}
	\toprule
    \textbf{Dataset} & \textbf{Acc} & \textbf{P} & \textbf{R} & \textbf{F1}\\
    \midrule
    AskParents & 83.71 & 76.86 & 79.62 & 73.14 \\
    \midrule
    needadvice & 85.99 & 85.71 & 79.99 & 79.55 \\
    \bottomrule
\end{tabular}

    \caption{Average inter-annotator agreement for all workers against DS labels}
    \label{tab:dataset-agreement}
\end{table}

\paragraph{Agreement} Table~\ref{tab:dataset-agreement} evaluates the agreement between annotators in terms of  micro-averaged accuracy, precision, recall and F1 between each worker and the DS labels. These numbers, although moderately high, show that there is  disagreement among workers. However, \citet{nye-etal-2018-corpus} found that despite the internal noise with complex tasks, the aggregated labels can still align well with experts. Table~\ref{tab:dataset-agreement} also shows that agreement scores are higher on r/needadvice than on r/AskParents. 

Table~\ref{tab:dataset-internal} reports the Kappa values of the resolved expert labels against either the DS labels or majority vote. We find that DS labels have substantial agreement with expert labels, and that the agreement is higher than majority vote. This result confirms that the aggregated DS labels are reliable. 

\paragraph{A note on posts with deleted question bodies} We observed after collecting annotations that 69 of 407 posts in r/AskParents and 98 of 277 posts in r/needadvice had been deleted by users or removed by moderators, meaning the submission bodies were missing and only the titles and comment-trees remained. However, most of the titles of these question posts are highly informative, and provide ample context for advice annotation, as shown below:

\ex. How can I enjoy my loneliness? \label{ex:deleted-1}

\ex. If I quit a grocery store job after two shifts, will I have to report it for employement history? \label{ex:deleted-2}

We identified 19 deleted posts whose titles failed to provide annotators with enough context. However, since we found no discrepancy with the the agreement scores for any annotations from these posts, we don't exclude them from the dataset. We report the agreement scores within deleted posts for both subreddits in Table~\ref{tab:agreement-deleted} in the Appendix.

\subsection{Corpus} Our final dataset consists of annotations of 407 posts in r/AskParents (by 95 workers) and 277 posts in r/needadvice (by 64 workers). Table~\ref{tab:dataset-sents} gives an overview of the sentence metrics in our dataset, along with the fraction of sentences DS-labeled as advice. We used a train/development/test split of 80-10-10 on posts rather than sentences so as to retain context for sentences in the same post.

\begin{table}[t]
    \centering
    \footnotesize
    \begin{tabular}{llll}
	\toprule
    \textbf{Dataset} & \textbf{Train} & \textbf{Dev} & \textbf{Test} \\
    \midrule
    AskParents & 8701(.29) & 802(.33) & 1091(.26)\\
    \midrule
    needadvice & 6148(.37) & 816(.34) & 898(.37) \\
    \bottomrule
\end{tabular}

    \caption{Sentence metrics in our dataset, with fraction DS-labeled as advice.}
    \label{tab:dataset-sents}
\end{table}

\section{Preliminary Analysis}
\label{sec:prelimanalysis}
\subsection{How is advice expressed?}

As noted previously, r/AskParents and r/needadvice differ with respect to their styles of moderation, but they are also different communities that may engage in giving advice differently. To understand how this impacts the structure of replies to posts, we manually analyzed 10 different posts from r/AskParents, and 4 different posts each from the flairs of r/needadvice.

We observed that people often give advice by alluding to their personal experience, for example:

\ex. I did the classic Ferberizing : check on baby after 5 mins , then 10 mins , then 20 mins , etc , until asleep . \label{ex:discourse-1}

Otherwise, a range of pragmatic strategies are adopted as noted by \citet{abolfathiasl2013pragmatic}, including the use of questions, imperatives, conditionals, etc.:

\ex. Have you tried a calm spray ? \label{ex:discourse-2}

\ex. Figure out why they like them , and then recommend those ones for those reasons . \label{ex:discourse-3}

\ex. If he does n’t want therapy , maybe an antidepressant would help . \label{ex:discourse-4}

Personal narratives are particularly interesting because it can be used to express advice indirectly, as in example (2). Table \ref{tab:preliminary-analysis-modes} reports the percentage of advice sentences that contain personal narratives. We analyzed 213 sentences DS-labelled as advice from 13 posts for whether they contained personal narratives. We observe that r/AskParents has a higher percentage (16.4\%) of personal narrative sentences than r/needadvice overall (6.33\%), though \textit{Mental Health} posts in r/needadvice have a high percentage of sentences that expressed personal narratives, at 18.18\%. These statistics, as well as the lower agreement statistics for r/AskParents which we report in Table \ref{tab:dataset-agreement}, suggest that r/AskParents is in general a harder dataset to work with.

\begin{table}[t]
    \centering
    \footnotesize
    \begin{tabular}{lll}
	\toprule
    \textbf{Subreddit} & \textbf{Other (\%)} & \textbf{Personal}\\
     &  & \textbf{Narrative (\%)}\\
    \midrule
    \textbf{r/AskParents} & 83.6 & 16.4\\
    \midrule
    \textbf{r/needadvice} & 93.67 & 6.33 \\
    \textbf{\quad-Career} & 100 & 0\\
    \textbf{\quad-Mental Health} & 81.82 & 18.18\\
    \textbf{\quad-Friendships} & 100 & 0\\
    \textbf{\quad-Education} & 95.4 & 4.6\\
    \textbf{\quad-Life Decisions} & 88.9 & 11.1\\
    \bottomrule
\end{tabular}

    \caption{Modes of discourse for advice sentences in each flair/subreddit}
    \label{tab:preliminary-analysis-modes}
    \vspace{-4mm}
\end{table}

Personal narrative versus other advice-giving strategies demonstrates distinctions in \emph{discourse modes} of advice. \citet{smith_2003} recognizes 5 different discourse modes -- narrative, descriptive, report, information and argumentative -- which roughly identify a text's contribution through clusters of linguistic features including temporal progression, stative vs.\ generic sentences, etc. We found that personal narrative is often expressed in the \emph{narrative} discourse mode, as shown in example \ref{ex:discourse-1} above. For non-personal-narrative advice, the \emph{argumentative} discourse mode is highly prevalent, as shown in example \ref{ex:discourse-3} above. Additionally, we have also observed the \emph{information} discourse mode, where the advice-giver expresses known facts in a general stative:

\ex. Just a bit of female health advice, having a late period is very normal \label{ex:discourse-5}

Finally, we noticed that advice-givers will tend to hedge their advice towards the end with a condition or possible consequences of following their advice, or as a form of reassurance. Take the following example from our dataset:

\ex. \textbf{Q}: Help. Accidentally fed one month old 4oz of baby water... Will she be okay?  \textbf{A}: She will absolutely be fine . Water is n't bad for a baby , \emph{though} obviously formula / breast milk is best.edit : You 're a good mom for being concerned \emph{though} .

\begin{figure}[t]
    \centering
    \includegraphics[width=\columnwidth]{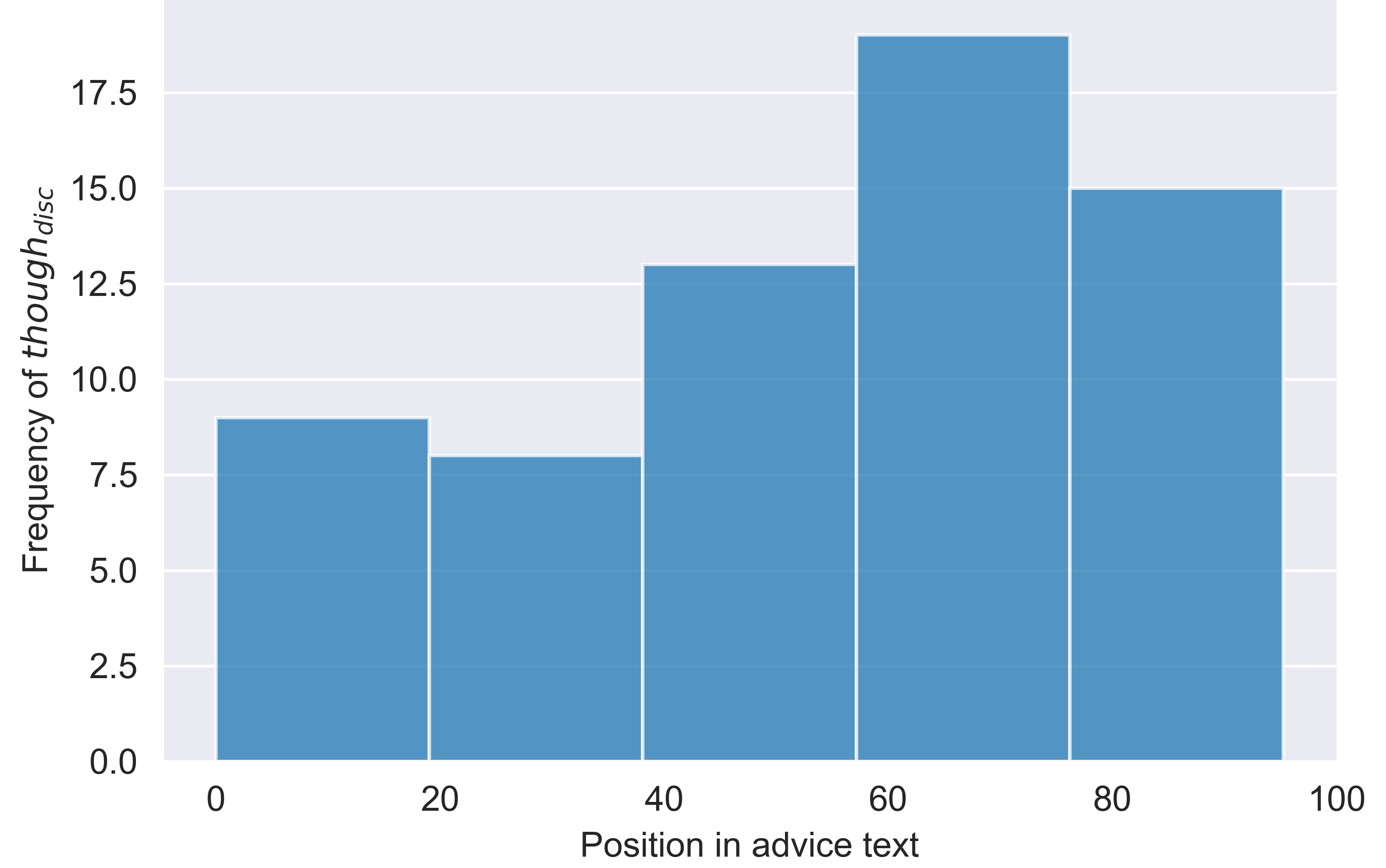}
    \caption{Frequency of discourse connective \emph{though}. X-axis: Frequency, Y-axis: Percentage progress through a reply, 0 is beginning and 100 is end of reply.}
    \label{fig:edu-though}
    \vspace{2mm}
\end{figure}

The discourse marker ``though'' is frequently used for signalling concession and contrast~\citep{pdtb}. This intuition is confirmed by an analysis of the discourse connective ``though'' among all posts we collected, which revealed a clear tendency towards the end of a reply, as illustrated in Figure \ref{fig:edu-though}.The lexical discourse marker ``though'' was found by splitting a large collection of posts and replies from r/AskParents into Elementary Discourse Units \citep{mann1988rhetorical}, using a neural discourse segmenter \cite{wang-etal-2018-toward}. 

\subsection{Non-advice sentences in advice posts}
\label{subsec:nonadvice}

Table~\ref{tab:dataset-sents} shows that the majority of sentences in replies to an advice-seeking post do not actually contain advice. To understand this phenomenon, we looked into sentences that are annotated as non-advice in our dataset. We found several distinctive phenomena, some of which are described with examples below (non-advice text is italicised):

\ex. Expressing sentiment: I also found being fully prepared for an interview calmed me down \ldots \emph{Good luck on your interviews and fingers crossed .} 

\ex. Providing support to advice: Look for smaller outfits , they ’re more likely to be willing to give you some time . \emph{Most professionals - if they have the time - are more than happy to talk to a student about what they do , especially if the student is interested in the same field .}

\ex. Reasoning about the situation: \emph{Yes , no one will ever know the big answers to the big questions .} What is the only thing that if shared , will grow larger in size?Answer : Love . Let that define your actions in life .

These non-advice sentences suggest a highly dynamic way in which advice-giving is structured into a coherent discourse. They also indicate that context can play a role in identifying advice.

\subsection{Lexical Analysis}
\label{subsec:lexicalanalysis}

\begin{table}[t]
    \centering
    \small
    \begin{tabular}{lp{30mm}p{30mm}}
	\toprule
     & \textbf{Advice} & \textbf{Non-advice} \\
    \midrule
    {\multirow{8}{*}{\rotatebox{90}{r/AskParents}}} & book if take something help then you might talk need down can etc play find show or great also give buy big watch diaper car about else minute spend baby & luck sorry shit however dog crazy teenager op die eventually three wish weird daughter yeah brother example miss gender anyway anymore comment morning lol boyfriend girl younger hope drive mine\\
    \midrule
    {\multirow{9}{*}{\rotatebox{90}{r/needadvice}}}
    & he phone night adult stay set big game doctor fun bring less show love depend activity eat normal put teacher family etc minute teach allow home they area & luck degree company college interview hobby student field mental course op sorry job dog anxiety hire eventually position path shit comment human online community shoe thanks note exercise depression slowly\\
    \bottomrule
\end{tabular}

    \caption{Top 30 lemmas ranked by logodds ratio}
    \label{tab:logodds}
\end{table}

To motivate that the language of advice varies systematically from non-advice, we quantify how strongly individual lemmas are associated with advice versus non-advice text. We use the log-odds ratio as a metric of comparison \citep{nye-nenkova-2015-identification}. To counteract the tendency of log-odds scores to highlight infrequent lemmas \citep{monroe_colaresi_quinn_2017}, we filter out lemmas that occurred less than 20 times in the train and validation set of our corpus. 

Table \ref{tab:logodds} shows the top 30 lemmas (excluding punctuation characters and numbers) from advice and non-advice sentences for each subreddit ranked by their log-odds ratio. We observe that there are fewer verbs among non-advice lemmas than advice lemmas, and that lemmas which are generally used in expressing sentiment (\textit{luck}, \textit{sorry}, \textit{thanks}) are more likely to be found in non-advice sentences. Combined with our observations in \S\ref{subsec:nonadvice}, this shows that language varies systematically between advice and non-advice sentences.

\section{Models}
\label{sec:models}
\paragraph{Task setup} We have constructed a dataset from the subreddits r/needadvice and r/AskParents as a general purpose resource for studying the breadth of advice-giving strategies. Our modelling experiments aim to establish baseline performance for rule-based models and language models at identifying advice, as well as explore how their performance varies with domain and provided context. We model advice identification as a binary classification task -- given a sentence, predict whether the sentence is advice or not.

\paragraph{Baselines} We test the baseline rule-based model and the top performing rule-based submission \citep[\textbf{NTUA-IS};][]{potamias-etal-2019-ntua} from SEMEVAL Task 9 2019 on our dataset, and use the results of these rule based models as baselines against which to gauge the performance of more advanced ones based on pre-trained 
language models. 

The baseline model provided by \citet{negi-etal-2019-semeval} uses search patterns to identify suggestions, including words (\emph{suggest, recommend}), phrases (\emph{.*would\textbackslash slike.*if.*}), and part-of-speech (POS) tags (\emph{modals, past tense verbs}).

However, some of these rules are naive and not intepretable -- such as classifying a sentence as a suggestion if it  contains a modal or the base form of a verb. \citet{potamias-etal-2019-ntua} improve upon this baseline with more keywords and phrases, searching for more rigorous POS patterns within clauses rather than sentences, and assigning different confidence scores for keyword and POS matches\footnote{Due to the lack of availability of code from \citet{potamias-etal-2019-ntua}, we attempted to reverse engineer all of their rules to the best of our ability.}. A sentence is classified as a suggestion if it exceeds a preset confidence score.

Since there is broad overlap between the purposes of their task and our analysis, we believe the results of these rule-based models are good baselines for our dataset. Moreover, the lexical and linguistic rules provide avenues of analysis for interpreting how our models make predictions.

\paragraph{Utilizing pre-trained language models} Pre-trained language models based on the Transformer architecture \citep{vaswani_attention_2017} subsequently finetuned on a dataset relevant to the downstream task of interest have proven to be immensely successful in NLP. Therefore, we consider two model architectures based on BERT~\citep{devlin_bert:_2019}. We finetune models separately on r/AskParents and r/needadvice.

BERT has been pretrained for classification tasks with a  special \textsc{[cls]} token appended at the beginning of  the sentence. We use this token's final hidden layer  representation exclusively for classification. We experiment  with 3 different ways of passing inputs to the pre-trained language model, varying the presence of some form of context:

\begin{enumerate}
    \item \textbf{BERT$_{\text{sent}}$}: We only use the 
    sentence as input.
    \item \textbf{BERT$_{\text{sent+q}}$}: BERT has also 
    been pretrained for question-answering tasks with a
    \textsc{cls} token followed by two spans of text with a 
    separation (\textsc{[sep]}) token between them, like so: 
    \textsc{[cls] sentence a [sep] sentence b}. We set 
    \textsc{sentence A} as the sentence being classified and 
    \textsc{sentence B} as the title and last three 
    sentences of the corresponding advice-seeking post.
    \item \textbf{BERT$_{\text{sent+c}}$}: In addition to
    using the advice-seeking post as context for the sentence, we experiment with
    using the rest of the reply as context. We set \textsc{sentence b} as the
    remainder of the reply by that user.
\end{enumerate}

We also present results for non-finetuned BERT embeddings (\textbf{BERT$_{\text{noft}}$}), where we only finetune the parameters of the classifier on top of the BERT model.

\paragraph{Generalizability} We explore the generalizability of models finetuned on  r/AskParents and r/needadvice by taking the best performing model on each dataset and analyze the predictions of the model on the other dataset. Since our r/AskParents dataset is larger, we also experiment with training on a subset of r/AskParents that is similar in size to r/needadvice.

\paragraph{Implementation} We use the \emph{bert-base-cased} pretrained embeddings from HuggingFace's Transformers module \citep{Wolf2019HuggingFacesTS}.  All models are optimized with AdamW \citep{loshchilov2018decoupled} and fine tuned for a maximum of 6 epochs with early stopping. We used a batch size of 32, and set weight decay to 0 and learning rate to \texttt{1e-5}.

\paragraph{Evaluation} We report precision, and recall and F1 scores for all models. The results for the finetuned BERT-based models are averaged over 5 random restarts during finetuning, and presented along with their standard deviation in parentheses.

\section{Results}
\label{sec:results}
\begin{table}[t]
    \centering
    \small
    \begin{tabular}{cllll}
	\toprule
    & \textbf{Model} & \textbf{P} & \textbf{R} & \textbf{F1} \\
    \midrule
    {\multirow{6}{*}{\rotatebox{90}{\tiny r/AskParents}}} & SEMEVAL & 32.7 & 70.2 & 44.6 \\
    & NTUA-IS & 31.4 & 64.9 & 42.3 \\
    & BERT$_{\text{noft}}$ & 62.6 (1.2) & 14.9 (1.0) & 24.0 (1.4) \\
    & BERT$_{\text{sent}}$ & 54.9 (2.4) & 49.5 (4.4) & \textbf{51.9} (1.9) \\
    & BERT$_{\text{sent+c}}$ & 54.2 (2.1) & 49.9 (4.0) & 51.9 (2.2) \\
    & BERT$_{\text{sent+q}}$ & 61.0 (13.4) & 33.1 (11.9) & 37.4 (8.1) \\
   \midrule
   {\multirow{6}{*}{\rotatebox{90}{\tiny r/needadvice}}} & SEMEVAL & 44.5 & 80.3 & 57.2 \\
    & NTUA-IS & 43.0 & 70.9 & 53.5 \\
    & BERT$_{\text{noft}}$ & 82.9 (0.5) & 44.6 (1.4) & 58.0 (1.2) \\
    & BERT$_{\text{sent}}$ & 79.7 (3.8) & 76.3 (3.9) & \textbf{77.8} (0.3) \\
    & BERT$_{\text{sent+c}}$ & 80.4 (4.4) & 75.3 (4.4) & 77.6 (0.7) \\
    & BERT$_{\text{sent+q}}$ & 83.4 (4.8) & 64.7 (7.4) & 72.5 (3.5) \\
    \bottomrule
\end{tabular}
    \caption{Classification results on test set.}
    \label{tab:results-classification}
\end{table}

\paragraph{Baseline} The performance of the baseline models and the finetuned language models are given in Table~\ref{tab:results-classification}. Surprisingly, we find that our baseline rule-based models perform reasonably well -- they outperform non-finetuned BERT embeddings at recall. However, as noted previously, many of the keyword and POS pattern rules are simplistic, which explains their high false positive rate.

\begin{table}[t]
    \centering
    \small
    \begin{tabular}{lccc}
	\toprule
    \textbf{Model} & \textbf{P} & \textbf{R} & \textbf{F1} \\
    \midrule
     AP $\to$ AP & 54.9 (2.4) & 49.5 (4.4) & 51.9 (1.9) \\
     AP$_{p}$ $\to$ AP & 59.1 (3.5) & 44.4 (4.1) & 50.5 (1.8) \\
     NA $\to$ AP & 61.9 (4.9) & 39.7 (3.5) & 48.1 (1.3) \\
    \midrule
     NA $\to$ NA & 79.7 (3.8) & 76.3 (3.9) & 77.8 (0.3) \\
     AP $\to$ NA & 74.0 (4.0) & 79.3 (2.9) & 76.5 (0.9) \\
     AP$_{p}$ $\to$ NA & 76.9 (3.8) & 75.5 (4.7) & 76.0 (1.1) \\
    \bottomrule
\end{tabular}
    \caption{Generalizbility results on test set. AP=r/AskParents, AP$_p$ = AP subset, NA =r/needadvice}
    \label{tab:results-transfer}
\end{table}

\paragraph{r/AskParents vs r/needadvice} We observe that all of the models perform better on the r/needadvice dataset, providing further evidence that r/AskParents is a more challenging dataset.  As already discussed, this is likely due to a combination of factors -- r/AskParents is less moderated than r/needadvice, and contains a higher proportion of narrative compared to argumentative discourse modes.

\paragraph{BERT$_{\text{sent+c}}$} We observe that adding context to a post does not improve model performance. This could be because the architecture we used to add context to the model, \textsc{[cls] sent [sep] context [sep]}, may not be conducive to retrieving contextual information necessary to identify advice. 

\paragraph{BERT$_{\text{sent+q}}$} Curiously, appending information from the question using the same architecture leads to a noticeable loss in model performance along with high variability. This could be because the question and the sentence are written by different users, leading to discourse incoherence which might confuse the model. For instance, while BERT$_{\text{sent}}$ classified the following sentence correctly, appending the question title and last 3 sentences of the question body lead it to go astray:

\ex. \textbf{Sentence}:You don't actually have to tell her anything of any substance. \textbf{Question}: Why is my Mother so negative over my new job? The end Rant over, thank you all

We experimented with only appending the question title, as well as excluding posts that had deleted post bodies, and found similar loss in performance along with variability.

We have illustrated that context from the question (like in \ref{ex:advice-2}) and from the rest of the reply (like those in \S\ref{subsec:nonadvice}) can help in identifying advice. However, neither of our models with context outperforms the model without context. Future work needs to work on building better models that can extract relevant information from these contextual cues to inform advice identification.

\paragraph{Generalizability} Table~\ref{tab:results-transfer} shows that while testing on another advice domain leads to lower performance on both subreddit datasets, the model trained on r/AskParents, a more niche subreddit, performs well on the more general r/needadvice subreddit. Our model results suggest that data from both subreddits is sufficiently generalizable for models to learn some general features of what constitutes advice. Moreover, training on a subset of the r/AskParents data (71\% randomly sampled) doesn't lead to substantial degradation of performance on r/AskParents (or r/needadvice). This result indicates that models find it harder to learn from our r/AskParents dataset, since more data doesn't seem to lead to substantial improvements in performance.

\paragraph{Flairs} Table \ref{tab:results-flair} reports per-flair results (of the BERT$_{\text{sent}}$ model) on r/needadvice. We observe that the lowest performance is in the flairs Mental Health and Career. We had shown previously (Table~\ref{tab:preliminary-analysis-modes}) that Mental Health had a high proportion of personal narrative discourse, which we can see tends to lead to lower performance. For Career, the reasons are less clear.

\begin{table}[t]
    \centering
    \small
    \begin{tabular}{llll}
	\toprule
    \textbf{Flair} & \textbf{P} & \textbf{R} & \textbf{F1} \\
    \midrule
    Friendships & 85.5 (5.7) & 93.8 (0.0) & 89.2 (2.9) \\
    Mental Health & 75.6 (3.5) & 74.7 (3.6) & 75.0 (0.6) \\
    Education & 86.8 (2.9) & 67.4 (6.2) & 75.7 (3.1) \\
    Career & 75.9 (5.1) & 78.0 (3.8) & 76.7 (1.3)\\
    Life Decisions & 82.4 (4.4) & 82.8 (3.5) & 82.4 (0.7) \\
    \bottomrule
\end{tabular}
    \caption{Flair results on test set.}
    \label{tab:results-flair}
\end{table}

\section{Analysis}
\label{sec:analysis}
We chose the BERT$_{\text{sent}}$ model -- the best performing model on both datasets, and analyzed the attention weights to see if they show some of the patterns we used in the baseline models. The attention weights were visualized using BertViz \citep{vig2019transformervis}.

\begin{figure}[t]
    \centering
    \includegraphics[width=0.8\columnwidth]{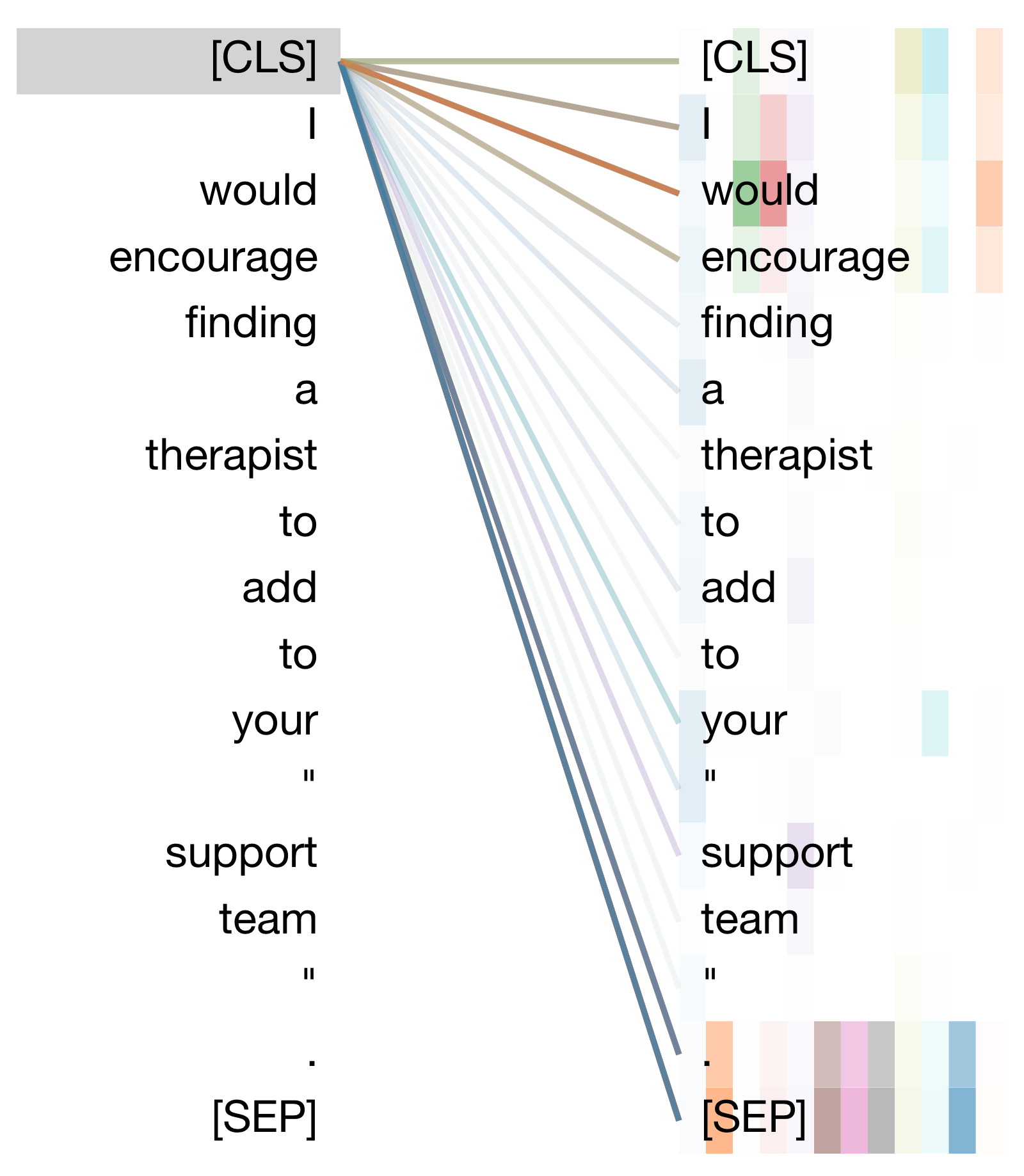}
    \caption{Attention distribution of a reply to a post titled \textit{Parenting with a History of
             Depression?}.}
    \label{fig:att-1}
\end{figure}

\paragraph{Attention Analysis} Transformer based language models utilize multiple self-attention heads to learn higher order and long distance relationships among words in a sentence. In Figure \ref{fig:att-1}, we visualize the distribution of attention weights from the final hidden layer, with each color representing a different attention head. The \textsc{[cls]} token is observed to attend to the modals  that the baseline rule based models have explicitly encoded in them. 

\begin{figure}[t]
    \centering
    \includegraphics[width=0.8\columnwidth]{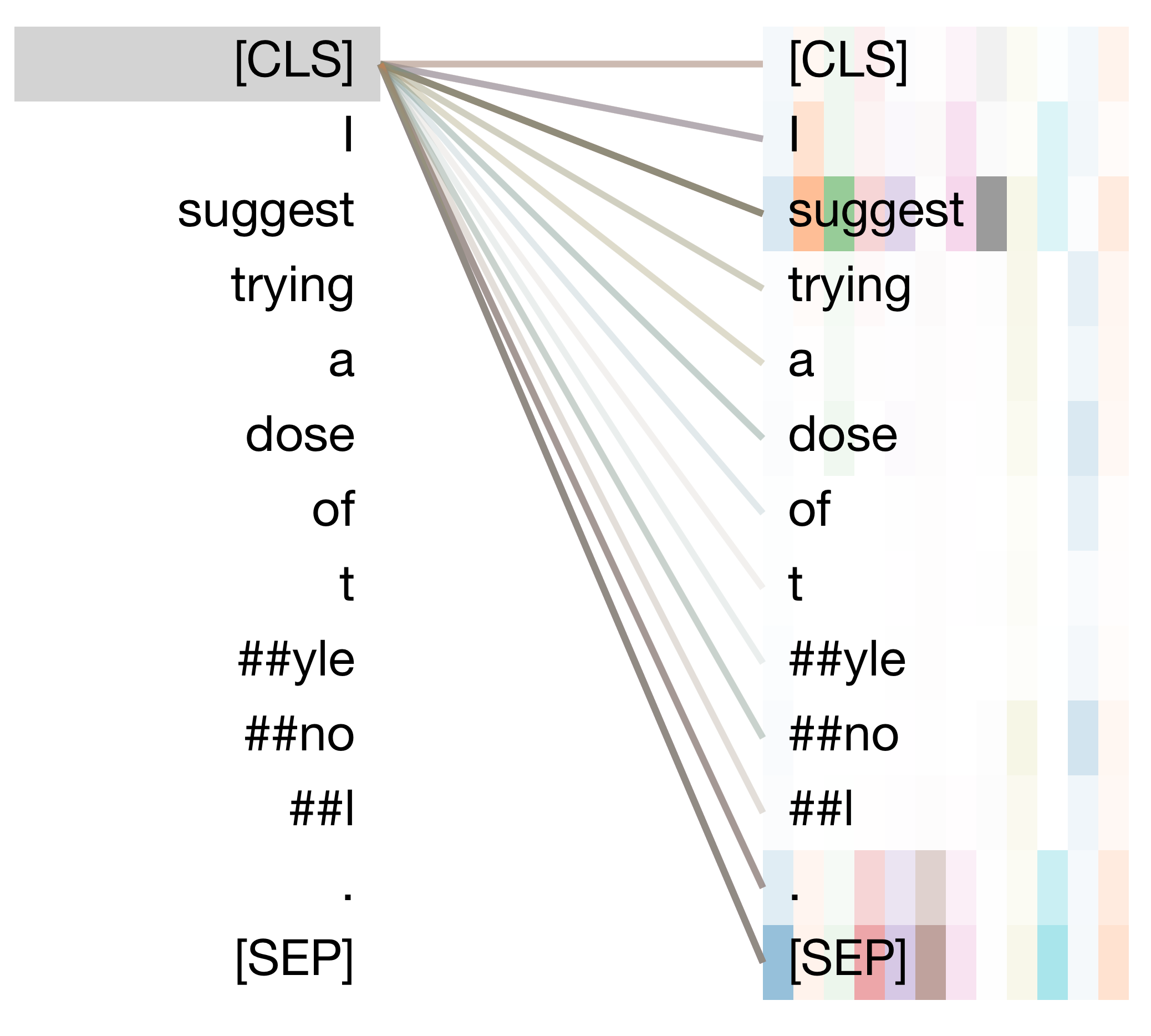}
    \caption{Attention distribution of a reply to a post titled \textit{Why will my 10 month old not stop crying?}.}
    \label{fig:att-2}
\end{figure}

The model is also robust to noise in our annotation protocol. The sentence in Figure \ref{fig:att-2}, was improperly annotated as not advice, as was the aggregated DS label. However in Figure \ref{fig:att-2}, which visualizes the attention distribution in the penultimate layer, we observe that the model attends to \textit{suggest}, and correctly predicts this sentence to contain advice. This is promising, since it shows that finetuned language models are latching onto surface level syntactic and lexical cues that we know to be indicative of advice.

\paragraph{Narrative Discourse} Narrative discourse is known to contain higher instances of advice that is given implicitly~\cite{abolfathiasl2013pragmatic}. For instance, the following is a different reply to the same post dicussed in Figure \ref{fig:att-1}:

\ex. I talked on Reddit with others to get support and ideas .

The user is implicitly suggesting to the advice-seeker that they should talk  with others on Reddit, since it helped them. This span was annotated as advice, but our model predicts otherwise. To understand if the model struggles with personal narratives, we analysed its performance on sentences that contain the personal pronouns \textit{me}, \textit{my} or \textit{we} which we take as indicative of personal narrative. A cursory analysis of the validation sets found 109 such sentences in r/AskParents, 81 of which we consider to be personal narratives, and 100 such sentences in r/needadvice, 66 of which we consider to be personal narratives.

Table \ref{tab:analysis-personal} shows that the model performance suffers on sentences that are approximated to contain personal narratives. We also observe a higher variability in the performance of the models, which indicates that the model is also highly uncertain of its predictions in such contexts. Future work on advice identification needs to look into how this can 
be improved using discourse level information.

\begin{table}[t]
    \centering
    \small
    \begin{tabular}{llll}
	\toprule
    \textbf{Dataset} & \textbf{P} & \textbf{R} & \textbf{F} \\
    \midrule
    AP & 54.9 (2.4) & 49.5 (4.4) & 51.9 (1.9) \\
    AP$_{\text{pers}}$ & 43.4 (4.3) & 31.7 (5.7) & 32.2 (7.7) \\
    NA & 79.7 (3.8) & 76.3 (3.9) & 77.8 (0.3) \\
    NA$_{\text{pers}}$ & 61.2 (16.3) & 37.9 (6.9) & 45.9 (6.9) \\
    \bottomrule
\end{tabular}
    \caption{Performance of model on test set comprising only personal narrative sentences. AP=r/AskParents,  NA=r/needadvice}
    \label{tab:analysis-personal}
    \vspace{-4mm}
\end{table}

\section{Conclusion}
\label{sec:conclusion}
We introduce a new dataset on advice given on the online platform Reddit, specifically r/AskParents and r/needadvice that differ in audience and level of moderation. We find that advice language consists of various pragmatic strategies and discourse structures. 
We find that fine-tuned BERT discovers certain surface-level features indicative of advice, but struggles to disambiguate instances of implicit advice conveyed  through personal narrative. Future work needs to look into how question and reply context can improve automatic identification of advice.

\section*{Acknowledgments}
We thank the anonymous reviewers for their valuable feedback. We are grateful to family and friends who supported the authors personally during the COVID-19 pandemic. This work was partially supported by a Salesforce Deep Learning Research Grant and NSF Grant IIS-1850153. We acknowledge the \href{http://www.tacc.utexas.edu}{Texas Advanced Computing Center (TACC)} at The University of Texas at Austin for providing HPC resources that have contributed to the research results reported within this paper.

\bibliography{references}
\bibliographystyle{acl_natbib}

\clearpage
\appendix
\section*{Appendix}
\label{sec:appendix}
\begin{table}[H]
    \centering
    \begin{tabular}{cllll}
	\toprule
    & \textbf{Model} & \textbf{P} & \textbf{R} & \textbf{F1} \\
    \midrule
    {\multirow{6}{*}{\rotatebox{90}{\tiny AskParents}}} & SEMEVAL & 38.27 & 67.54 & 48.85 \\
    & NTUA-IS & 36.49 & 60.45 & 45.51 \\
    & BERT$_{\text{noft}}$ & 74.20 (1.61) & 22.68 (0.77) & 34.74 (0.77) \\
    & BERT$_{\text{sent}}$ & 62.93 (3.36) & 58.95 (4.69) & 60.70 (1.84) \\
    & BERT$_{\text{sent+c}}$ & 61.84 (2.68) & 61.64 (4.72) & 61.59 (1.89) \\
    & BERT$_{\text{sent+q}}$ & 66.41 (9.80) & 46.55 (10.20) & 53.46 (4.11) \\
   \midrule
   {\multirow{6}{*}{\rotatebox{90}{\tiny NeedAdvice}}} & SEMEVAL & 42.01 & 82.48 & 55.67 \\
    & NTUA-IS & 37.23 & 68.61 & 48.27 \\
    & BERT$_{\text{noft}}$ & 74.72 (0.30) & 43.80 (1.06) & 55.22 (0.89) \\
    & BERT$_{\text{sent}}$ & 68.76 (2.98) & 73.72 (4.65) & 71.00 (0.90) \\
    & BERT$_{\text{sent+c}}$ & 71.23 (3.29) & 71.97 (5.09) & 71.41 (1.23) \\
    & BERT$_{\text{sent+q}}$ & 73.19 (1.70) & 61.17 (9.75) & 66.21 (5.48) \\
    \bottomrule
\end{tabular}
    \caption{Classification results on validation set.}
    \label{tab:results-classification-dev}
\end{table}

\begin{table}[H]
    \centering
    \begin{tabular}{lccc}
	\toprule
    \textbf{Model} & \textbf{P} & \textbf{R} & \textbf{F1} \\
    \midrule
     AP $\to$ AP & 62.93 (3.36) & 58.95 (4.69) & 60.70 (1.84) \\
     AP$_{p}$ $\to$ AP & 66.76 (3.87) & 53.28 (6.05) & 58.94 (2.36) \\
     NA $\to$ AP & 68.02 (5.49) & 51.19 (6.37) & 57.95 (2.52) \\
    \midrule
     NA $\to$ NA & 68.76 (2.98) & 73.72 (4.65) & 71.00 (0.90) \\
     AP $\to$ NA & 58.68 (2.77) & 80.29 (4.71) & 67.68 (1.29) \\
     AP$_{p}$ $\to$ NA & 67.73 (3.44) & 70.51 (4.69) & 68.91 (0.89) \\
    \bottomrule
\end{tabular}
    \caption{Generalizability results on validation set.}
    \label{tab:results-transfer-dev}
\end{table}

\begin{table}[H]
    \centering
    \begin{tabular}{llllll}
	\toprule
    \textbf{Dataset} & \textbf{Acc} & \textbf{P} & \textbf{R} & \textbf{F1}\\
    \midrule
    r/AskParents(D) & 86.18 & 79.46 & 74.7 & 72.89 \\

    r/AskParents(ND) & 83.22 & 76.38 & 80.54 & 73.21 \\

    r/needadvice(D) & 87.21 & 85.21 & 81.03 & 79.48 \\

    r/needadvice(ND) & 85.38 & 85.96 & 79.48 & 79.58 \\
    \bottomrule
\end{tabular}

    \caption{IAA on deleted(D) and not-deleted(ND) posts against DS labels.}
    \label{tab:agreement-deleted}
\end{table}

\begin{table}[H]
    \centering
    \begin{tabular}{llll}
	\toprule
    \textbf{Dataset} & \textbf{Train} & \textbf{Dev} & \textbf{Test} \\
    \midrule
    AskParents & 327 & 40 & 40\\
    \midrule
    needadvice & 223 & 27 & 27 \\
    \bottomrule
\end{tabular}
    \caption{Post-level metrics on our dataset.}
    \label{tab:dataset-posts}
\end{table}

\end{document}